\documentclass{article}
\usepackage{ismir,amsmath,cite,url}
\usepackage{graphicx}
\usepackage{color}

\usepackage{xspace}
\usepackage{bbm}
\usepackage{url}
\usepackage{amsfonts}
\usepackage{algpseudocode}
\usepackage{algorithm}
\usepackage{subfigure}

\title{Counterpoint by Convolution}
\multauthor
{Cheng-Zhi Anna Huang$^{\star\dagger\sharp}$ \hspace{1cm}
 Tim Cooijmans$^{\star\dagger}$ \hspace{1cm}
 Adam Roberts$^\sharp$}
{ \bfseries{
 Aaron Courville$^\dagger$ \hspace{1cm}
 Douglas Eck$^\sharp$}\\
 $^\star$ Equal contributions \hspace{.5cm}
 $^\dagger$ MILA, Universit\'e de Montr\'eal \hspace{.5cm}
 $^\sharp$ Google Brain
 \\
{\tt\small chengzhiannahuang@gmail.com, cooijmans.tim@umontreal.ca}
}
\def\authorname{Cheng-Zhi Anna Huang, Tim Cooijmans, Adam Roberts, Aaron Courville, Douglas Eck}

\newcommand{\vect}[1]{\mathbf{#1}}
\newcommand{\given}{~|~}
\newcommand{\expect}{\mathbb{E}}
\newcommand{\ctx}{C}
\newcommand{\coconet}{\textsc{Coconet}\xspace}

\newcommand{\nade}{\textsc{Nade}\xspace}

\begin{document}

\maketitle
%
%
%
\begin{abstract}
Machine learning models of music typically break up the task of composition into a chronological process, composing a piece of music in a single pass from beginning to end.
On the contrary, human composers write music in a nonlinear fashion, scribbling motifs here and there, often revisiting choices previously made.
In order to better approximate this process, we train a convolutional neural network to complete partial musical scores, and explore the use of blocked Gibbs sampling as an analogue to rewriting.
Neither the model nor the generative procedure are tied to a particular causal direction of composition.

Our model is an instance of orderless \nade~\cite{uria2014deep}, which allows more direct ancestral sampling.
However, we find that Gibbs sampling greatly improves sample quality, which we demonstrate to be due to some conditional distributions being poorly modeled.
Moreover, we show that even the cheap approximate blocked Gibbs procedure from~\cite{yao2014equivalence} yields better samples than ancestral sampling, based on both log-likelihood and human evaluation.
\end{abstract}

\section{Introduction}\label{sec:introduction}

\begin{figure}[h!]
\begin{center}
\includegraphics[trim=2.6cm 0.5cm 2.6cm 0, clip=true, width=1\linewidth]{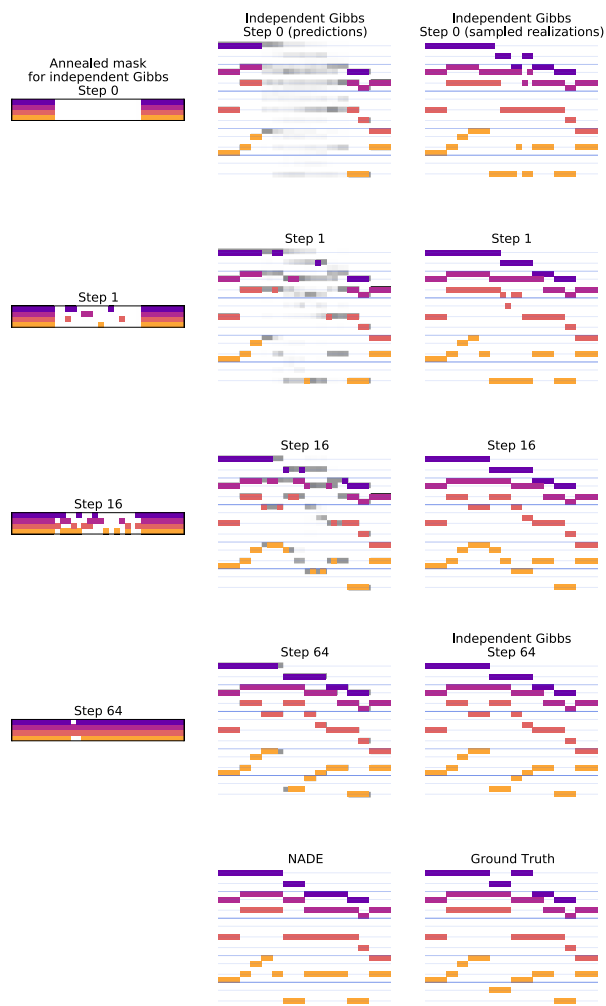}
\end{center}
\caption{
Blocked Gibbs inpainting of a corrupted Bach chorale by \coconet.
At each step, a random subset of notes is removed, and the model is asked to infer their values.
New values are sampled from the probability distribution put out by the model, and the process is repeated.
Left: annealed masks show resampled variables. Colors distinguish the four voices.
Middle: grayscale heatmaps show predictions $p(\vect{x}_j \given \vect{x}_\ctx)$ summed across instruments.
Right: complete pianorolls after resampling the masked variables. 
Bottom: a sample from NADE (left) and the original Bach chorale fragment (right).
}
\label{fig:process}
\end{figure}

Counterpoint is the process of placing notes against notes to construct a polyphonic musical piece.~\cite{fux1965study}  This is a challenging task, as each note has strong musical influences on its neighbors and notes beyond.
  Human composers have developed systems of rules to guide their compositional decisions.
However, these rules sometimes contradict each other, 
and can fail to prevent their users from going down musical dead ends.
Statistical models of music, which is our current focus, is one of the many computational approaches that can help composers try out ideas more quickly, thus reducing the cost of exploration~\cite{fernandez2013ai}.

Whereas previous work in statistical music modeling has relied mainly on sequence models such as
Hidden Markov Models (HMMs~\cite{baum1966statistical}) and
Recurrent Neural Networks (RNNs~\cite{rumelhart1988learning}),
we instead employ convolutional neural networks due to  their invariance properties and emphasis on capturing local structure.
Nevertheless, they have also been shown to successfully model large-scale structure~\cite{oord2016pixel,oord2016wavenet}.
Moreover, convolutional neural networks have shown to be extremely versatile once trained, as demonstrated by a variety of creative uses such as DeepDream~\cite{mordvintsev2015inceptionism} and style transfer~\cite{gatys2015neural}.

We introduce \coconet, a deep convolutional model trained to reconstruct partial scores.
Once trained, \coconet provides direct access to all conditionals of the form
$p(\vect{x}_i \given \vect{x}_\ctx)$ where
$\ctx$ selects a fragment of a musical score $\vect{x}$ and $i \notin \ctx$ is in its complement.  \coconet is an instance of deep orderless \nade~\cite{uria2014deep},
which learns an ensemble of factorizations of the joint $p(\vect{x})$, each corresponding to a different ordering.  
A related approach is the multi-prediction training of deep Boltzmann machines (MP-DBM)~\cite{goodfellow2013multi}, which also gives a model that can predict any subset of variables given its complement.

However, the sampling procedure for orderless \nade treats the ensemble as a mixture and relies heavily on ordering.
Sampling from an orderless \nade involves (randomly) choosing an ordering,
and sampling variables one by one according to the chosen ordering.
This process is called \emph{ancestral sampling}, as the order of sampling follows the directed structure of the model.
%
%
%
%
%
%
%
%
We have found that this produces poor results for the highly structured and complex domain of musical counterpoint.
%
%
%
%

Instead, we propose to use blocked-Gibbs sampling, a Markov Chain Monte Carlo method to sample from a joint probability distribution by repeatedly resampling subsets of variables using conditional distributions derived from the joint probability distribution.
An instance of this was previously explored by~\cite{yao2014equivalence} who employed a \nade in the transition operator for a Markov Chain, yielding a Generative Stochastic Network (GSN).
The transition consists of a corruption process that masks out a subset $\vect{x}_{\neg\ctx}$ of variables,
followed by a process that independently resamples variables $\vect{x}_i$ (with $i \notin \ctx$) according to the distribution $p_\theta(\vect{x}_i \given \vect{x}_\ctx)$ emitted by the model with parameters $\theta$.
%
%
%
%
%
%
%
%
%
%
Crucially, the effects of independent sampling are amortized by annealing the probability with which variables are masked out.
Whereas~\cite{yao2014equivalence} treat their procedure as a cheap approximation to ancestral sampling,
we find that it produces superior samples.
Intuitively, the resampling process allows the model to iteratively \emph{rewrite} the score, giving it the opportunity to correct its own mistakes.

\coconet addresses the general task of completing partial scores; special cases of this task include "bridging" two musical fragments, and temporal upsampling and extrapolation.
Figure~\ref{fig:process} shows an example of \coconet populating a partial piano roll using blocked-Gibbs sampling. Code and samples are publically available.\footnote{\begin{tabular}{ l l }
   Code:& \tiny \url{https://github.com/czhuang/coconet} \\
   Data:&  \tiny \url{https://github.com/czhuang/JSB-Chorales-dataset} \\
  Samples:& \tiny \url{https://coconets.github.io/} \\

\end{tabular}
}
Our samples on a variety of generative tasks such as rewriting, melodic harmonization and unconditioned polyphonic music generation show the versatility of our model.
In this work we focus on Bach chorales, and assume four voices are active at all times.
However, our model can be easily adapted to the more general, arbitrarily polyphonic representation as used in~\cite{boulanger2012modeling}.
%
%
%
%


Section~\ref{sec:related} discusses related work in modeling music composition, with a focus on counterpoint.
The details of our model and training procedure are laid out in Section~\ref{sec:model}.
%
%
We discuss evaluation under the model in Section~\ref{sec:evaluation}, and sampling from the model in Section~\ref{sec:sampling}.
Results of quantitative and qualitative evaluations are reported in Section~\ref{sec:experiments}.  
%
\section{Related Work} \label{sec:related}
Computers have been used since their early days for experiments in music composition.  A notable composition is Hiller and Issacson's string quartet Illiac Suite~\cite{hiller1957musical}, which experiments with statistical sequence models such as Markov chains.  One challenge in adapting such models is that music consists of multiple interdependent streams of events.  Compare this to typical sequence domains such as speech and language, which involve modeling a single stream of events: a single speaker or a single stream of words.  In music, extensive theories in counterpoint have been developed to address the challenge of composing multiple streams of notes that coordinate.  One notable theory is due to Fux~\cite{fux1965study} from the Baroque period, which introduces \emph{species counterpoint} as a pedagogical scheme to gradually introduce students to the complexity of counterpoint.   In first species counterpoint only one note is composed against every note in a given fixed melody (\emph{cantus firmus}), with all notes bearing equal durations and the resulting vertical intervals consisting of only consonances.    
%
%

Computer music researchers have taken inspiration from this pedagogical scheme by first teaching computers to write species counterpoint as opposed to full-fledged counterpoint.  Farbood~\cite{farbood2001analysis} uses Markov chains to capture transition probabilities of different melodic and harmonic transitions rules.  Herremans~\cite{herremans2012composing, herremans2013composing} takes an optimization approach by writing down an objective function that consists of existing rules of counterpoint and using a variable neighbourhood search (VNS) algorithm to optimize it.

J.S. Bach chorales has been the main corpus in computer music that serves as a starting point to tackle full-fledged counterpoint.  A wide range of approaches have been used to generate music in the style of Bach chorales, for example rule-based and instance-based approaches such as Cope's recombinancy method~\cite{cope1991computers}.  This method involves first segmenting existing Bach chorales into smaller chunks based on music theory, analyzing their function and stylistic signatures and then re-concatenating the chunks into new coherent works.
Other approaches range from  constraint-based~\cite{pachet2001musical} to statistical methods~\cite{conklin2003music}.  In addition, \cite{fernandez2013ai} gives a comprehensive survey of AI methods used not just for generating Bach chorales, but also algorithmic composition in general.  



Sequence models such as HMMs and RNNs are natural choices for modeling music.  
Successful application of such models to polyphonic music often requires serializing or otherwise re-representing the music to fit the sequence paradigm.
For instance, Liang in BachBot~\cite{liang2016bachbot} serializes four-part Bach chorales by interleaving the parts, while Allan and Williams~\cite{allan2005harmonising} construct a chord vocabulary.
Boulanger et al.~\cite{boulanger2012modeling} adopt a piano roll representation, a binary matrix $\vect{x}$ where $\vect{x}_{it} = 1$ iff some instrument is playing pitch $i$ at time $t$.
To model the joint probability distribution of the multi-hot pitch vector $\vect{x}_t$, they employ a Restricted Boltzmann Machine (RBM~\cite{smolensky1986information,hinton2006fast}) or Neural Autoregressive Distribution Estimator (NADE~\cite{larochelle2011neural}) at each time step. 
Similarly Goel et al.~\cite{goel2014polyphonic} employ a Deep Belief Network~\cite{hinton2006fast} on top of an RNN.


Hadjeres et al.~\cite{hadjeres2016style} instead employ an undirected Markov model to learn pairwise relationships between neighboring notes up to a specified number of steps away in a score.  
Sampling involves Markov Chain Monte Carlo (\textsc{MCMC}) using the model as a Metropolis-Hastings (\textsc{MH}) objective.  The model permits constraints on the state space to support tasks such as melody harmonization. 
However, the Markov assumption can limit the expressivity of the model. 

Hadjeres and Pachet in DeepBach~\cite{hadjeres2016deepbach} model note predictions by breaking down its full context into three parts, with the past and the future modeled by stacked LSTMs going in the forward and backward directions respectively, and the present harmonic context modeled by a third neural network.  The three are then combined by a fourth neural network and used in Gibbs sampling for generation.  

Lattner et al. imposes higher-level structure by interleaving selective Gibbs sampling on a convolutional RBM~\cite{lattner2016imposing} and gradient descent that minimizes cost to template piece on features  such as self-similarity.  This procedure itself is wrapped in simulated annealing  to ensure steps do not lower the solution quality too much. 

We opt for an orderless \nade training procedure which enables us to train a mixture of all possible directed models simultaneously. Finally, an approximate blocked Gibbs sampling procedure~\cite{yao2014equivalence} allows fast generation from the model. 

%
%
%
%




\section{Model} \label{conv}
\label{sec:model}
We employ machine learning techniques to obtain a generative model of musical counterpoint in the form of piano rolls.
%
%
%
%
%
%
Given a dataset of observed musical pieces $\vect{x}^{(1)} \ldots \vect{x}^{(n)}$ posited to come from some true distribution $p(\vect{x})$,
we introduce a model $p_\theta(\vect{x})$ with parameters $\theta$.
In order to make $p_\theta(\vect{x})$ close to $p(\vect{x})$, we maximize the data log-likelihood $\sum_i \log p_\theta(\vect{x}^{(i)})$ (an approximation of $\mathbb{E}_{\vect{x} \sim p(\vect{x})} \log p_\theta(\vect{x})$) by stochastic gradient descent.

The joint distribution $p(\vect{x})$ over $D$ variables $\vect{x}_1 \ldots \vect{x}_D$ is often difficult to model directly and hence we construct our model $p_\theta(\vect{x})$ from simpler factors.
In the \nade~\cite{larochelle2011neural} framework, the joint $p_\theta(\vect{x})$ is factorized autoregressively, one variable at a time, according to some ordering $o = o_1 \ldots o_D$,
%
%
%
%
%
%
%
%
such that
\begin{align}
\label{eqn:nadelikelihood}
p_\theta(\vect{x}) = \prod_d p_\theta(\vect{x}_{o_d} \given \vect{x}_{o_{<d}}).
\end{align}
For example, it can be factorized in chronological order:
 \begin{align}
p_\theta(\vect{x}) = p_\theta(\vect{x}_1) p_\theta(\vect{x}_2 | \vect{x}_1) \ldots p_\theta(\vect{x}_D | \vect{x}_{D-1} \ldots \vect{x}_1)
 \end{align}
 In general, \nade permits any one fixed ordering, and although all orderings are equivalent from a theoretical perspective,
they differ in practice due to effects of optimization and approximation.
%
%
%
%

Instead, we can train \nade for all orderings $o$ simultaneously using the orderless \nade~\cite{uria2014deep} training procedure.
This procedure relies on the observation that, thanks to parameter sharing, computing
$p_\theta(\vect{x}_{o_{d'}} \given \vect{x}_{o_{<d}})$ for all $d' \geq d$ is no more expensive than computing it only for $d' = d$.\footnote{Here $\vect{x}_{o_{<d}}$ is used as shorthand for variables $\vect{x}_{o_1} \ldots \vect{x}_{o_{d-1}}$ that occur earlier in the ordering.}
Hence for a given $o$ and $d$ we can simultaneously obtain partial losses for all orderings that agree with $o$ up to $d$:
\begin{align}
\label{eqn:onadeloss}
\mathcal{L}(\vect{x}; o_{<d}, \theta)
   &= - \sum_{o_d} \log p_\theta(\vect{x}_{o_d} \given \vect{x}_{o_{<d}}, o_{<d}, o_d)
\end{align}An orderless \nade model offers direct access to all distributions of the form $p_\theta(\vect{x}_i | \vect{x}_\ctx)$
conditioned on any set of contextual variables $\vect{x}_\ctx = \vect{x}_{o_{<d}}$ that might already be known.
%
%
%
%
This gives us a very flexible generative model; in particular, we can use these conditional distributions to complete arbitrarily partial musical scores.

To train the model, we sample a training example $\vect{x}$ and context $\ctx$ such that $|\ctx| \sim U(1, D)$,
and update $\theta$ based on the gradient of the loss given by Equation~\ref{eqn:onadeloss}.
This loss consists of $D - d + 1$ terms, each of which corresponds to one ordering.
To ensure all orderings are trained evenly we must reweight the gradients by $1 / (D - d + 1)$.
This correction, due to~\cite{uria2014deep}, ensures consistent estimation of the joint negative log-likelihood $\log p_\theta(\vect{x})$.

In this work, the model $p_\theta(x)$ is implemented by a deep convolutional neural network~\cite{krizhevsky2012imagenet}.
%
%
%
%
This choice is motivated by the locality of contrapuntal rules and their near-invariance to translation, both in time and in pitch space.
%
%
%
%
%
%
%
%

We represent the music as a stack of piano rolls encoded in a binary three-dimensional tensor $\vect{x} \in \{0, 1\}^{I \times T \times P}$.
%
%
%
%
Here $I$ denotes the number of instruments, $T$ the number of time steps, $P$ the number of pitches,
and $\vect{x}_{i,t,p} = 1$ iff the $i$th instrument plays pitch $p$ at time $t$.
We will assume each instrument plays exactly one pitch at a time, that is, $\sum_p \vect{x}_{i,t,p} = 1$ for all $i, t$.

Our focus is on four-part Bach chorales as used in prior work~\cite{allan2005harmonising,boulanger2012modeling,goel2014polyphonic,liang2016bachbot,hadjeres2016style}.
Hence we assume $I = 4$ throughout.
We constrain ourselves to only the range that appears in our training data (MIDI pitches 36 through 88).
%
%
%
%
Time is discretized at the level of 16th notes for similar reasons.
To curb memory requirements, we enforce $T = 128$ by randomly cropping the training examples.
%
%
%
%

Given a training example $\vect{x} \sim p(\vect{x})$,
we present the model with values of only a strict subset of its elements
$\vect{x}_\ctx = \{\vect{x}_{(i,t)} \given (i, t) \in \ctx\}$
and ask it to reconstruct its complement $\vect{x}_{\neg \ctx}$.
The input $\vect{h}^0 \in \{0,1\}^{2I \times T \times P}$ is obtained by masking the piano rolls $\vect{x}$
to obtain the context $\vect{x}_\ctx$
and concatenating this with the corresponding mask:
\begin{align}
\vect{h}^{0}_{i,t,p} &= \mathbbm{1}_{(i, t) \in \ctx} \vect{x}_{i,t,p} \\
\vect{h}^{0}_{I + i,t,p} &= \mathbbm{1}_{(i, t) \in \ctx}
\end{align}

where the time and pitch dimensions are treated as spatial dimensions to convolve over. Each instrument's piano roll $\vect{h}^0_i$ and mask $\vect{h^0_{I + i}}$ is treated as a separate channel and convolved independently.
%
%
%
%
%
%

With the exception of the first and final layers, all convolutions preserve the size of the hidden representation.
%
%
%
%
%
%
That is, we use ``same'' padding throughout and all activations have the same number of channels $H$, such that  $\vect{h}^l \in \mathbb{R}^{H \times T \times P}$ for all $1 < l < L$.
Throughout our experiments we used $L = 64$ layers and $H = 128$ channels.
After each convolution we apply batch normalization~\cite{ioffe2015batch} (denoted by $\mathrm{BN}( \cdot )$) with statistics tied across time and pitch.
Batch normalization rescales activations at each layer to have mean $\beta$ and standard deviation $\gamma$, which greatly improves optimization.
%
%
After every second convolution, we introduce a skip connection from the hidden state two levels below to reap the benefits of residual learning~\cite{he2015deep}.
\begin{align}
\vect{a}^l &= \mathrm{BN}(\vect{W}^l * \vect{h}^{l-1}; \gamma^l, \beta^l) \\
\vect{h}^l &=
           \left\{
	\begin{array}{ll}
\mathrm{ReLU}(\vect{a}^l + \vect{h}^{l-2}) \notag \\
    \qquad \quad \text{ if } 3<l<L-1 \text{ and } l \bmod 2 = 0 \notag \\
\mathrm{ReLU}(\vect{a}^l) \text{ otherwise} \notag 
	\end{array}
\right. \\
\vect{h}^L &= \vect{a}^L
\end{align}
The final activations $\vect{h}^L \in \mathbb{R}^{I \times T \times P}$ are passed through the softmax function to obtain predictions for the pitch at each instrument/time pair:
%
%
%
%
\begin{align}
\label{eqn:partial_loss}
p_\theta(\vect{x}_{i,t,p} \given \vect{x}_\ctx, \ctx) =
\frac{\exp(h^L_{i,t,p})}{\sum_p \exp(h^L_{i,t,p})}
\end{align}
The loss function from Equation~\ref{eqn:onadeloss} is then given by
%
%
%
%
%
%
%
\begin{align}
\label{eqn:loss}
\mathcal{L}(\vect{x}; \ctx, \theta)
   &= - \sum_{(i, t) \notin \ctx}
      \log p_\theta(\vect{x}_{i,t} \given \vect{x}_\ctx, \ctx)
\\ &= - \sum_{(i,t) \notin \ctx} \sum_p \vect{x}_{i,t,p}
%
%
      \log p_\theta(\vect{x}_{i,t,p} \given \vect{x}_\ctx, \ctx) \notag
\end{align}
where $p_\theta$ denotes the probability under the model with parameters
$\theta = \vect{W}^1, \gamma^1, \beta^1,\allowbreak
%
%
%
\ldots,\allowbreak
\vect{W}^{L-1}, \gamma^{L-1}, \beta^{L-1}$.  
We train the model by minimizing
\begin{align}
\label{eqn:expectedloss}
\expect_{\vect{x} \sim p(\vect{x})}
\expect_{\ctx \sim p(\ctx)}
\frac{1}{|\neg\ctx|}\mathcal{L}(\vect{x}; \ctx, \theta)
%
%
\end{align}
with respect to $\theta$ using stochastic gradient descent with step size determined by Adam~\cite{kingma2014adam}.
The expectations are estimated by sampling piano rolls $\vect{x}$ from the training set and drawing a single context $\ctx$ per sample.

%
%

\section{Evaluation} \label{sec:evaluation}

The log-likelihood of a given example is computed according to Algorithm~\ref{alg:eval} by repeated application of Equation~\ref{eqn:partial_loss}.
Evaluation occurs one frame at a time, within which the model conditions on its own predictions and does not see the ground truth.
Unlike notewise teacher-forcing, where the ground truth is injected after each prediction, the framewise 
evaluation is thus sensitive to accumulation of error.
This gives a more representative measure of quality of the generative model.
For each example, we repeat the evaluation process a number of times to average over multiple orderings, and finally average across frames and examples.
For chronological evaluation, we draw only orderings that have the $t_l$s in increasing order.

\begin{algorithm}
\caption{Framewise log-likelihood evaluation}
\label{alg:eval}
\begin{algorithmic}
\State Given a piano roll $\vect{x}$
\State $L_{m,i,t} \gets 0$ for all $m,i,t$
\For{multiple orderings $m = 0 \ldots M$}
  \State $\ctx \gets \emptyset$, $\widehat{\vect{x}} \gets \vect{x}$
  \State Sample an ordering $t_1, t_2 \ldots t_T$ over frames
  \For{$l = 0 \ldots T$}
    \State Sample an ordering $i_1, i_2 \ldots i_I$ over instruments
    \For{$k = 0 \ldots I$}
      \State $\pi_p \gets p_\theta(\vect{x}_{i_k,t_l,p} \given \widehat{\vect{x}}_\ctx, \ctx)$ for all $p$
      \State $L_{m,i_k,t_l} \gets \sum_p \vect{x}_{i_k,t_l,p} \log \pi_p$
      \State $\widehat{\vect{x}}_{i_k,t_l} \sim \operatorname{Cat}(P, \pi)$
      \State $\ctx \gets \ctx \cup {(i_k, t_l)}$
    \EndFor
    \State $\widehat{\vect{x}}_\ctx \gets \vect{x}_\ctx$
  \EndFor
\EndFor
\\ \Return $-\frac{1}{T} \sum_t \log \frac{1}{M} \sum_m \exp \sum_i L_{m,i,t}$ \\
\end{algorithmic}
\end{algorithm}

\begin{table*}

\label{tab:likelihood}
\begin{center}
\begin{tabular}{llll}
\multicolumn{1}{c}{\bf Model} & \multicolumn{3}{c}{\bf Temporal resolution} \\
 & \multicolumn{1}{c}{quarter} & \multicolumn{1}{c}{eighth} & \multicolumn{1}{c}{sixteenth} \\
\hline%
\noalign{\vskip .5ex}
\nade~\cite{boulanger2012modeling}                     & 7.19 \\
\textsc{RNN-RBM}~\cite{boulanger2012modeling}          & 6.27 \\
\textsc{RNN}-\nade~\cite{boulanger2012modeling}        & 5.56 \\
\hline%
\noalign{\vskip .5ex}
\textsc{RNN}-\nade (our implementation) & 5.03 & 3.78 & 2.05 \\
\coconet (chronological)  & $7.79 \pm 0.09$ & $4.21 \pm 0.05$ & $2.22 \pm 0.03$ \\
\coconet (random)         & $5.03 \pm 0.06$ & $1.84 \pm 0.02$ & $0.57 \pm 0.01$ \\
\end{tabular}
\end{center}
\caption{
Framewise negative log-likelihoods (NLLs) on the Bach corpus.
We compare against~\protect\cite{boulanger2012modeling}, who used quarter-note resolution.
We also compare on higher temporal resolutions (eighth notes, sixteenth notes), against our own reimplementation of \textsc{RNN}-\nade.
\coconet is an instance of orderless \nade, and as such we evaluate it on random orderings.
However, the baselines support only chronological frame ordering, and hence we evaluate our model in this setting as well.
}
\end{table*}

\section{Sampling} \label{sec:sampling}

We can sample from the model using the orderless \nade ancestral sampling procedure, in which we first sample an ordering and then sample variables one by one according to the ordering.
%
%
%
%
%
%
However, we find that this yields poor samples, and we propose instead to use Gibbs sampling.

%
%

\subsection{Orderless \nade Sampling} \label{sec:sampling-nade}
%
%
%
%

Sampling according to orderless \nade involves first randomly choosing an ordering and then sampling variables one by one according to the chosen ordering.
%
%
We use an equivalent procedure in which we arrive at a random ordering by at each step randomly choosing the next variable to sample.
We start with an empty (zero everywhere) piano roll $\vect{x}^0$
%
%
%
%
and empty context $\ctx^0$ and populate them iteratively by the following process.
%
%
%
%
We feed the piano roll $\vect{x}^s$ and context $\ctx^s$ into the model to obtain a set of categorical distributions $p_\theta(\vect{x}_{i,t} | \vect{x}^s_{\ctx^s}, \ctx^s)$ for $(i, t) \notin \ctx^s$.
As the $\vect{x}_{i,t}$ are not conditionally independent, we cannot simply sample from these distributions independently.
However, if we sample from one of them, we can compute new conditional distributions for the others.
Hence we randomly choose one $(i, t)^{s+1} \notin \ctx^s$ to sample from, and let $\vect{x}^{s+1}_{i,t}$ equal the one-hot realization.
Augment the context with $C^{s+1} = C^s \cup (i,t)^{s+1}$ and repeat until the piano roll is populated.
%
%
%
%
This procedure is easily generalized to tasks such as melody harmonization and partial score completion by starting with a nonempty piano roll.

Unfortunately, samples thus generated are of low quality,
which we surmise is due to accumulation of errors.
This is a well-known weakness of autoregressive models.~\cite{venkatraman2015improving,bengio2015scheduled,huszar2015not,lamb2016professor}
%
%
%
%
While the model provides conditionals $p_\theta(\vect{x}_{i, t} | \vect{x}_\ctx, \ctx)$
for all $(i, t) \notin \ctx$, some of these conditionals may be better modeled than others.
We suspect in particular those conditionals used early on in the procedure,
for which the context $\ctx$ consists of very few variables.
Moreover, although the model is trained to be order-agnostic,
different orderings invoke different distributions,
which is another indication that some conditionals are poorly learned.
We test this hypothesis in Section~\ref{sec:sample-likelihoods}.

\subsection{Gibbs Sampling} \label{sec:sampling-gibbs}

To remedy this, we allow the model to revisit its choices:
we repeatedly mask out some part of the piano roll and then repopulate it.
This is a form of blocked Gibbs sampling~\cite{liu1994collapsed}.
%
%
Blocked sampling is crucial for mixing, as the high temporal resolution
of our representation causes strong correlations between consecutive notes.
For instance, without blocked sampling, it would take many steps to snap out of a long-held note.
Similar considerations hold for the Ising model from statistical mechanics,
leading to the Swendsen-Wang algorithm~\cite{swendsen1987nonuniversal}
in which large clusters of variables are resampled at once.

We consider two strategies for resampling a given block of variables:
\emph{ancestral} sampling and \emph{independent} sampling.
Ancestral sampling invokes the orderless \nade sampling procedure described in Section~\ref{sec:sampling-nade} on the masked-out portion of the piano roll.
Independent sampling simply treats the masked-out variables $\vect{x}_{\neg\ctx}$ as independent given the context $\vect{x}_\ctx$.

Using independent blocked Gibbs to sample from a \nade model has been studied by~\cite{yao2014equivalence}, who propose to use an annealed masking probability $\alpha_n = \max(
\alpha_{\mathrm{min}},
\alpha_{\mathrm{max}}
- n(\alpha_{\mathrm{max}} - \alpha_{\mathrm{min}}) / (\eta N)
)$ for some minimum and maximum probabilities $\alpha_{\mathrm{min}}, \alpha_{\mathrm{max}}$, total number of Gibbs steps $N$ and fraction $\eta$ of time spent before settling onto the minimum probability $\alpha_{\mathrm{min}}$.
%
%
%
%
Initially, when the masking probability is high, the chain mixes fast but samples are poor due to independent sampling.
As $\alpha_n$ decreases, the blocked Gibbs process with independent resampling approaches standard Gibbs where one variable at a time is resampled, thus amortizing the effects of independent sampling.
%
%
%
%
%
%
%
%
%
%
$N$ is a hyperparameter which as a rule of thumb we set equal to $IT$; it can be set lower than that to save computation at a slight loss of sample quality.
%
%
%
%
%
%
%
%

\cite{yao2014equivalence} treat independent blocked Gibbs as a cheap approximation to ancestral sampling.
Whereas plain ancestral sampling (\ref{sec:sampling-nade}) requires $O(IT)$ model evaluations, ancestral blocked Gibbs requires a prohibitive $O(ITN)$ model evaluations and independent Gibbs requires only $O(N)$, where $N$ can be chosen to be less than $IT$.
%
%
%
%
%
%
%
%
Moreover, we find that independent blocked Gibbs sampling
in fact yields \emph{better} samples than plain ancestral sampling.
%
%

\section{Experiments}
\label{sec:experiments}

We evaluate our approach on a popular corpus of four-part Bach chorales.
While the literature features many variants of this dataset~\cite{allan2005harmonising,boulanger2012modeling,liang2016bachbot,hadjeres2016style}, we report results on that used by~\cite{boulanger2012modeling}.
As the quarter-note temporal resolution used by~\cite{boulanger2012modeling} is frankly too coarse to accurately convey counterpoint,
we also evaluate on eighth-note and sixteenth-note quantizations of the same data.

It should be noted that quantitative evaluation of generative models is fundamentally hard~\cite{theis2015note}.
The gold standard for evaluation is qualitative comparison by humans, and we therefore report human evaluation results as well.


\subsection{Data Log-likelihood}

Table~\ref{tab:likelihood} compares the framewise log-likelihood of the test data under variants of our model and those reported in~\cite{boulanger2012modeling}.
We find that the temporal resolution has a dramatic influence on the performance, which we suspect is an artifact of the performance metric. The log-likelihood is evaluated by teacher-forcing, that is, the prediction of a frame is conditioned on the ground truth of all previously predicted frames.
As temporal resolution increases, chord changes become increasingly rare, and the model is increasingly rewarded for simply holding notes over time. 

We evaluate \coconet on both chronological and random orderings, in both cases averaging likelihoods across an ensemble of $M=5$ orderings.
The chronological orderings differ only in the ordering of instruments within each frame.
We see in Table~\ref{tab:likelihood} that fully random orderings lead to significantly better performance.
We believe the members of the more diverse random ensemble are more mutually complementary.
For example, a forward ordering is uncertain at the beginning of a piece and more certain toward the end, whereas a backward ordering is more certain at the beginning and less certain toward the end.

\subsection{Sample Quality} \label{sec:sample-likelihoods}

In Section~\ref{sec:sampling} we conjectured that the low quality of \nade samples
is due to poorly modeled conditionals $p_\theta(\vect{x}_{i, t} \given \vect{x}_\ctx, \ctx)$
where $\ctx$ is small.
We test this hypothesis by evaluating the likelihood under the model
of samples generated by the ancestral blocked Gibbs procedure with $\ctx$
chosen according to independent Bernoulli variables.
When we set the inclusion probability $\rho$ to 0, we obtain \nade.
Increasing $\rho$ increases the expected context size $|\ctx|$,
which should yield better samples if our hypothesis is true.
The results shown in Table~\ref{tab:sample-likelihoods} confirm that this is the case. 
For these experiments, we used sample length $T = 32$ time steps and number of Gibbs steps $N = 100$.

\begin{table}[h]

\label{tab:sample-likelihoods}
\begin{center}
\begin{tabular}{lrr}
\multicolumn{1}{c}{\bf Sampling scheme} &\multicolumn{1}{c}{\bf Framewise NLL}
\\ \hline%
\noalign{\vskip .5ex}
Ancestral Gibbs, $\rho=0.00$ (\nade) & $1.09 \pm 0.06$ \\
Ancestral Gibbs, $\rho=0.05$ & $1.08 \pm 0.06$ \\
Ancestral Gibbs, $\rho=0.10$ & $0.97 \pm 0.05$ \\
Ancestral Gibbs, $\rho=0.25$ & $0.80 \pm 0.04$ \\
Ancestral Gibbs, $\rho=0.50$ & $0.74 \pm 0.04$ \\
Independent Gibbs~\cite{yao2014equivalence} & $0.52 \pm 0.01$
\end{tabular}
\end{center}
\caption{
Mean ($\pm$ SEM) NLL under model of unconditioned samples generated from model by various schemes.
}
\end{table}

\begin{figure}[h]
\begin{center}
\includegraphics[width=0.85\linewidth, trim={0 0.2cm 0 0.2cm},clip]{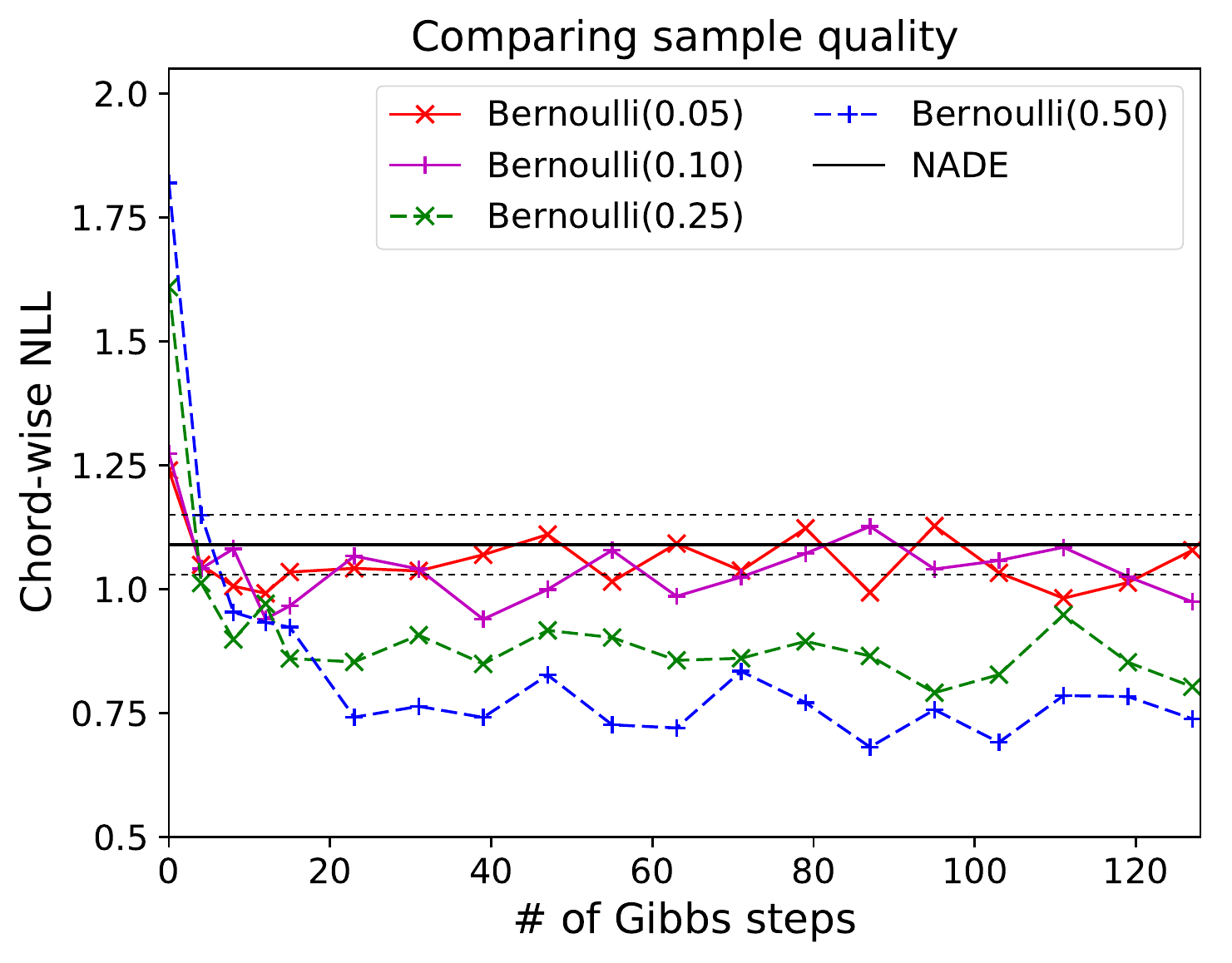}
\end{center}
\caption{
Likelihood under the model for ancestral Gibbs samples obtained with various context distributions $p(\ctx)$.
\nade ($\mathrm{Bernoulli(0.00)}$) is included for reference.
}
\label{fig:compare_nll}
\end{figure}

Figure~\ref{fig:compare_nll} shows the convergence behavior of the various Gibbs procedures, averaged over 100 runs.
We see that for low values of $\rho$ (small $\ctx$), the chains hardly make progress beyond \nade in terms of likelihood.
Higher values of $\rho$ (large $\ctx$)  enable the model to get off the ground and reach significantly better likelihood.
%
%
%
%
%
%

\subsection{Human Evaluations}
\label{sec:human-evaluations}
To further compare the sample quality of different sampling procedures, we carried out a listening test on Amazon's Mechanical Turk (MTurk).  The  procedures include orderless \nade ancestral sampling and
independent Gibbs~\cite{yao2014equivalence}, with each we generate four unconditioned samples of eight-measure lengths from empty piano rolls. 
To have an absolute reference for the quality of samples, we include first eight measures of four random Bach chorale pieces from the validation set.
Each fragment lasts thirty-four seconds after synthesis.

\begin{figure}
\begin{center}
\includegraphics[width=0.8\linewidth, trim={0 0.2cm 0 0.2cm},clip]{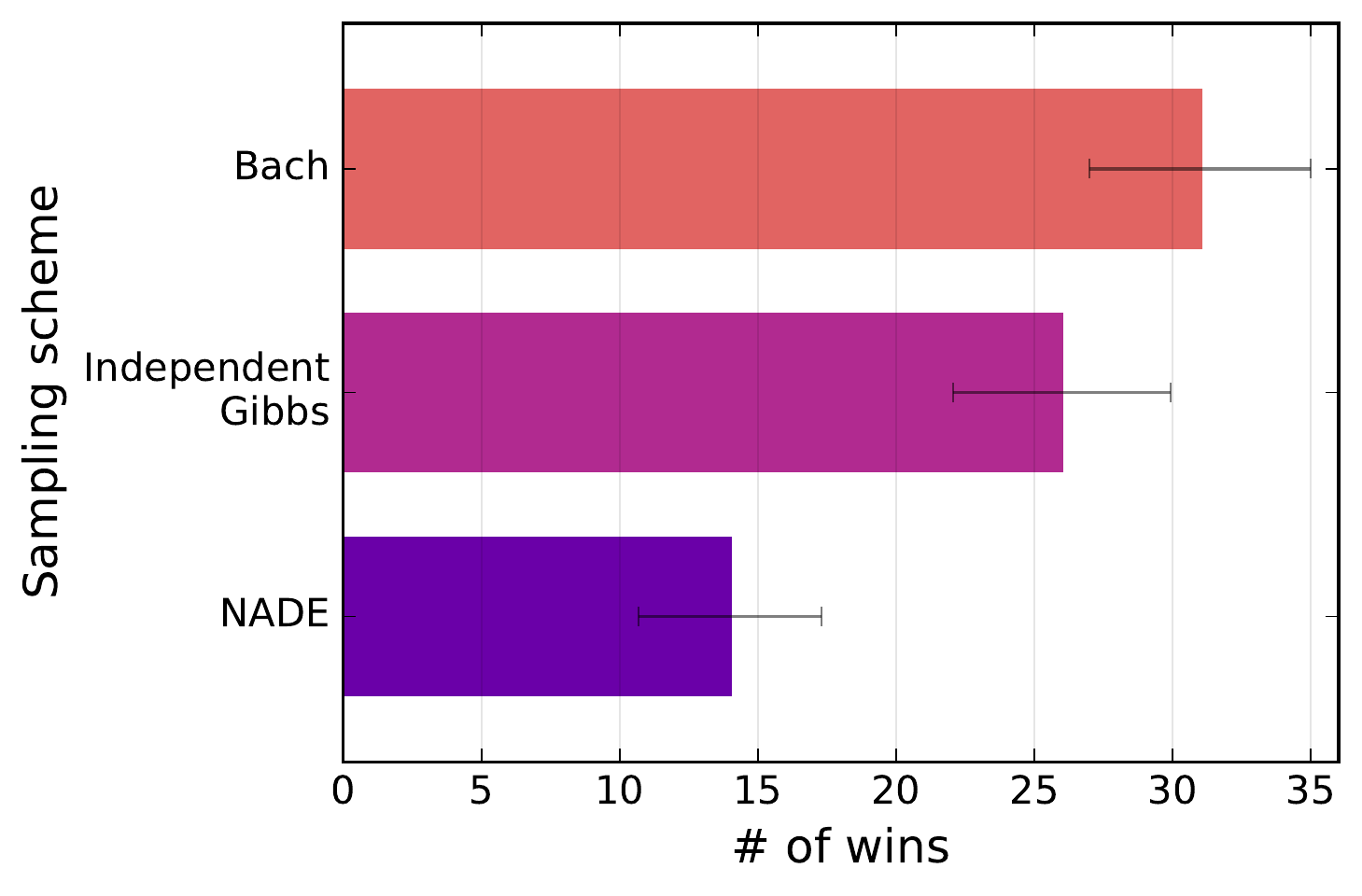}
\end{center}
\caption{
Human evaluations from MTurk on how many times a sampling procedure or Bach is perceived as more Bach-like.  Error bars show the standard deviation of a binomial distribution fitted to each's binary win/loss counts.}

\label{fig:unconditionedevaluation}
\end{figure}

For each MTurk hit, participants are asked to rate on a Likert scale which of the two random samples they perceive as more Bach-like.
A total of 96 ratings were collected, with each source involved in 64 (=96*2/3) pairwise comparisons.
Figure~\ref{fig:unconditionedevaluation} shows the number of times each source was perceived as closer to Bach's style. We perform a Kruskal-Wallis H test on the ratings, $\chi^2(2)=12.23, p<0.001$, showing there are statistically significant differences between models.  A post-hoc analysis using the Wilcoxon signed-rank test with Bonferroni correction showed that participants perceived samples from independent Gibbs as more Bach-like than ancestral sampling (\nade), $p<0.05/3$.  This confirms the loglikelihood comparisons on sample quality in~\ref{sec:sample-likelihoods} that independent Gibbs produces better samples.
There was also a significance difference between Bach and ancestral samples but not between Bach and independent Gibbs. 

\section{Conclusion} \label{sec:conclusion} 

We introduced a convolutional approach to modeling musical scores based on the orderless \nade~\cite{uria2016neural} framework.
Our experiments show that the \nade ancestral sampling procedure yields poor samples,
which we have argued is because some conditionals are not captured well by the model.
We have shown that sample quality improves significantly when we use blocked Gibbs sampling to iteratively rewrite parts of the score.
Moreover, annealed independent blocked Gibbs sampling as proposed by~\cite{yao2014equivalence}
is not only faster but in fact produces better samples.





\section*{Acknowledgments}

We thank Kyle Kastner, Guillaume Alain, Gabriel Huang, Curtis (Fjord) Hawthorne, the Google Brain Magenta team, as well as Jason Freidenfelds for helpful feedback, discussions, suggestions and support.
We also thank Calcul Qu\'ebec and Compute Canada for computational support.

\bibliography{coconet}

%
%
%
%

\end{document}